\documentclass[conference]{IEEEtran}
\IEEEoverridecommandlockouts
\usepackage{cite}
\usepackage{amsmath,amssymb,amsfonts}
\usepackage{algorithmic}
\usepackage{graphicx}
\usepackage{textcomp}
\usepackage{microtype}
\usepackage{xcolor}
\usepackage{url}
\usepackage{tikz}
\usepackage{pgfplots}
\usepackage{graphicx}
\usepackage{subfig}
\usepackage{bm}
\usepackage{microtype}
\usepackage{flushend}
\def\BibTeX{{\rm B\kern-.05em{\sc i\kern-.025em b}\kern-.08em
    T\kern-.1667em\lower.7ex\hbox{E}\kern-.125emX}}
    
\usepackage{todonotes} 

\begin{document}

\title{Flying~with~Cartographer:~Adapting~the~Cartographer 3D Graph SLAM Stack for UAV Navigation \\
\thanks{Authors are with University of Zagreb, Faculty of Electrical Engineering and Computing, Laboratory for Robotics and Intelligent Control Systems (LARICS), 
         10000 Zagreb, Croatia
        {\tt\small (juraj.orsulic, robert.milijas, ana.batinovic, lovro.markovic, antun.ivanovic, stjepan.bogdan) at fer.hr}}
}
\author{Juraj Oršulić, Robert Milijas, Ana Batinovic, Lovro Markovic, Antun Ivanovic, Stjepan Bogdan}

\maketitle

\begin{abstract}
This paper describes an application of the Cartographer graph SLAM stack as a pose sensor in a UAV feedback control loop, with certain application-specific changes in the SLAM stack such as smoothing of the optimized pose. Pose estimation is performed by fusing 3D LiDAR/IMU-based proprioception with GPS position measurements by means of pose graph optimisation. Moreover, partial environment maps built from the LiDAR data (\emph{submaps}) within the Cartographer SLAM stack are marshalled into OctoMap, an Octree-based voxel map implementation. The OctoMap is further used for navigation tasks such as path planning and obstacle avoidance.
\end{abstract}


\section{Introduction}
\label{sec:introduction}

We consider a closed feedback loop control system for autonomous aerial vehicle (UAV) navigation -- in our case, a quadcopter drone. In our setup, the vehicle uses a laser radar (LiDAR) for environment sensing and pose feedback, aided with an inertial sensor (IMU) and a satellite navigation receiver. These sensors form a proprioception subsystem whose localization solution is used as feedback input to the vehicle control algorithm.

More precisely, since the vehicle navigates a completely or partially unknown environment, the proprioception subsystem implements a \emph{simultaneous localization and mapping} (SLAM) algorithm. Therein, a map is produced which is also used for path planning and following.

The task of a SLAM implementation is to create a consistent world map from sensor observations. This task entails detecting previously visited areas and using this information to refine the trajectory (and its tail, the current vehicle pose estimate). In SLAM terminology, these events are called \emph{loop closures} and are key for reducing the amount of drift accumulated over time.

The focus of this paper is on \emph{Cartographer}, an open source LiDAR graph SLAM stack originally developed by Google \cite{Hess2016}, and its application in UAV control.

Architecture of a graph SLAM implementation 
is commonly viewed in literature as comprising two parts: the sensor-specific frontend, which processes the sensor data and constructs the pose graph, and the sensor-agnostic backend, which performs pose graph optimization.
Cartographer comprises two components: \emph{local trajectory builder} (also called local SLAM) and \emph{global SLAM}.

We will briefly discuss this architecture and how it relates to the frontend-backend division.
The local trajectory builder component is a part of the SLAM frontend. It uses a voxel occupancy grid scan matcher and map builder to build a set of temporally compact \emph{submaps}, wherein each scan is inserted into two adjacent submaps. The second part of the frontend is the \emph{constraint builder} subsystem of the global SLAM component, which uses a branch and bound-based \emph{fast correlative scan matcher} to discover loop closure correspondences between old submaps and new laser scans, and vice versa (new scans and old submaps). Each laser scan is associated with a trajectory node in the pose graph.

The global SLAM component implements the backend -- optimization of the pose graph using LMA non-linear least squares, and steers the search for loop closures. The trajectory is periodically re-optimized to include all newly discovered loop closure constraints, as well as GPS measurements, if available. As we will discuss in the following chapter, this can sometimes induce jumps in the pose feedback signal, causing problems in the vehicle control loop.

When using LiDARs for UAV navigation, the robotics community mostly relies on the lidar odometry and mapping (LOAM) algorithm \cite{Zhang_2014}. 
To the best of our knowledge, Cartographer has not been used for UAV navigation outside of our lab. This has motivated us to describe the changes we have made in the code base in order to be able to use it for reliable UAV navigation.

\section{Safe loop closures}
\label{sec:loops}

State of the art graph SLAM systems -- such as Cartographer \cite{Hess2016}, also \cite{Koide_2019}, and \cite{iSAM2} which is used as SLAM backend in \cite{Shan2018} and \cite{Ji2019} -- perform loop closures so that the estimated pose is momentarily swapped with the new estimate when a new optimized trajectory is available. This in turn causes a step signal to be present in the pose estimate, which can be detrimental if the pose estimate is fed back to a position control system.

\begin{figure}[tbp]
    \centering
    \includegraphics[width=\columnwidth,trim=1.5cm 0 2.0cm 1.6cm,clip]{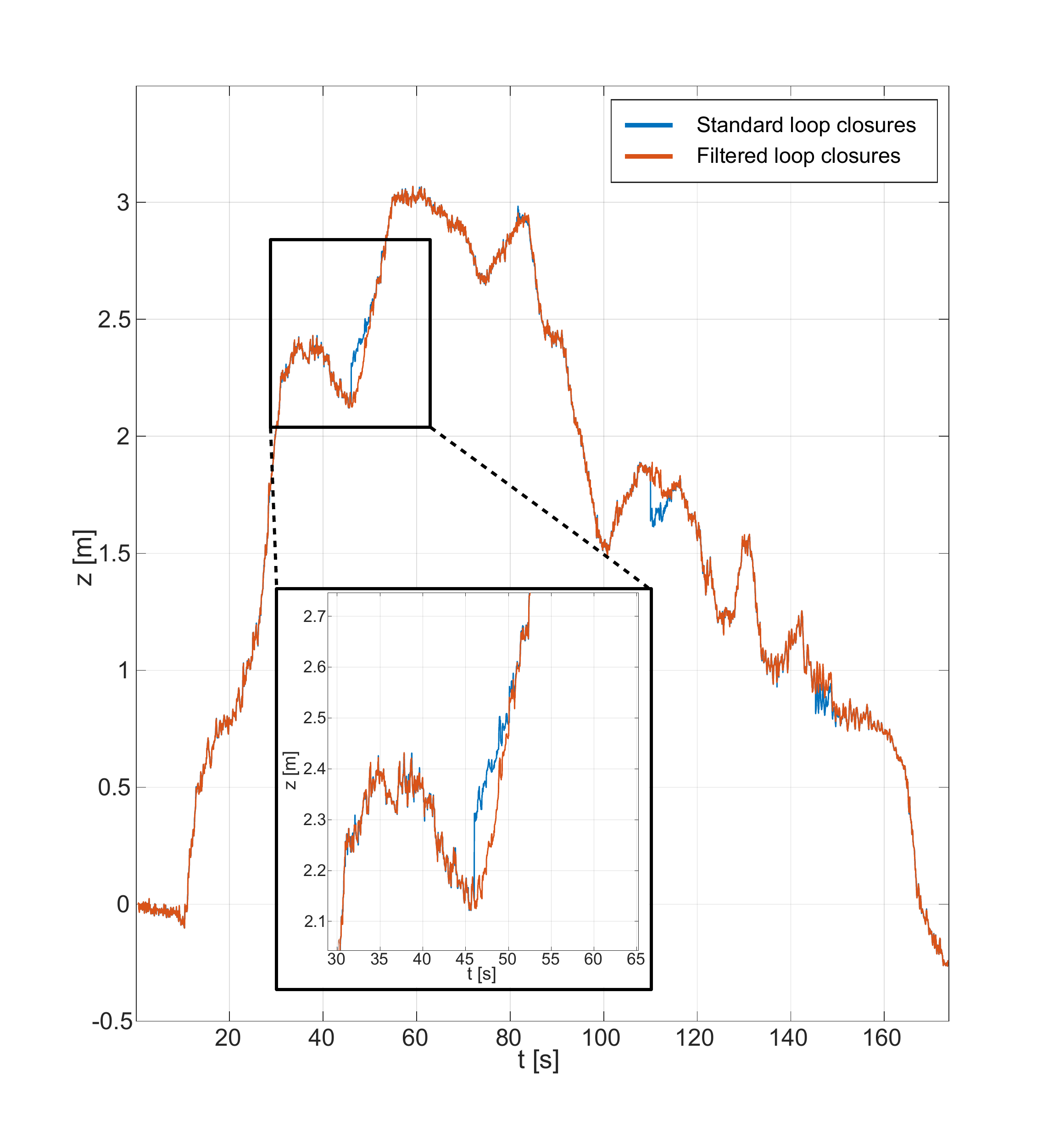}
    \caption{Comparison of the filtered pose estimate with the original Cartographer pose estimate in the event of a loop closure.}
    \label{fig:steps_in_feedback}
\end{figure}

Figure \ref{fig:steps_in_feedback} shows a typical SLAM pose estimate and the disturbances caused by pose graph optimization events after loop closures are discovered. From the perspective of UAV position control, this can be dangerous, since large disturbances in the pose feedback can cause the UAV controller to respond by performing an aggressive maneuver and crash into a nearby obstacle. 

To the best of our knowledge, there is no widely available graph SLAM implementation which closes loops in a safe manner for vehicle navigation. As a mitigation, we have added a hook onto the pose graph optimization event which starts a simple smoothing filter. The smoothing filter reduces the effect of the disturbance induced by trajectory optimization on the UAV position control system by smoothing the transition step across a seven-second window. This is enough for average UAV navigation use cases where large leaps due to loops being closed are not expected.

Cartographer computes the current global pose estimate as
\begin{equation}
    \label{eq:global_pose_calculation}
    \mathbf{T} = \mathbf{T}_\mathrm{map}^\mathrm{local} \; \cdot\; \mathbf{T}_\mathrm{local}^\mathrm{tracking}
\end{equation}
where $\mathbf{T}_\mathrm{local}^\mathrm{tracking}$ is the unoptimized vehicle pose without discontinuities as computed by local SLAM, while $\mathbf{T}_\mathrm{map}^\mathrm{local}$ contains the correction to the current ending of the local SLAM trajectory according to the most recent solution of pose graph optimization in global SLAM. When a new optimized solution is available, there is a sudden change of $\mathbf{T}_\mathrm{map}^\mathrm{local}$.

A trajectory optimization event triggers the start of smoothing ($t = 0$). Smoothing is performed by linear interpolation of $\mathbf{T}_\mathrm{map}^\mathrm{local}$ in a roughly seven-second window, during which the value of interpolation factor $\alpha$ takes on values in $[0, 1]$ according to
\begin{figure}[tbp]
    \centering
    \includegraphics[width=\columnwidth]{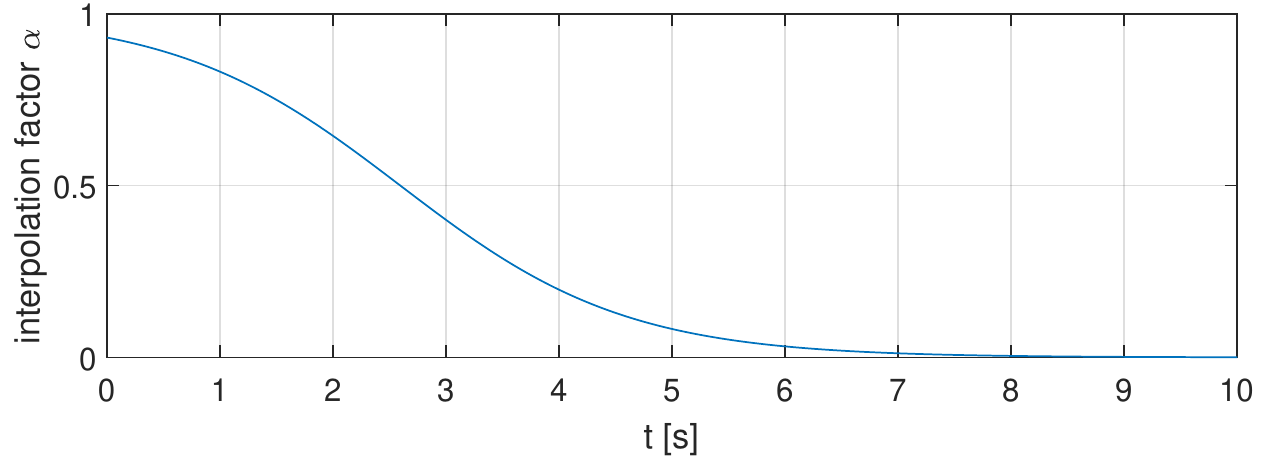}
    \caption{The interpolation factor function \eqref{eq:smooth}, a smooth transition from 1 to 0, reduces the disturbance of loop closure events in the vehicle control loop.}
    \label{fig:smoothing}
\end{figure}
\begin{align}
    \label{eq:smooth}
    \alpha(t) &= \frac{1}{1 + s \exp(t - t_x) }, \nonumber \\
    s &= 1.5, \quad t_x = 3 \; \mathrm{s}.
\end{align}
The numerical constants $s$ and $t_x$ were chosen empirically so that the transition is smooth in a roughly seven-second window after a new optimized solution is available (Figure \ref{fig:smoothing}). The translatory part $\mathbf{p}$ of the transformation $\mathbf{T}_\mathrm{map}^\mathrm{local}$ is interpolated as $\alpha \mathbf{p}_\mathrm{old} + (1 - \alpha) \mathbf{p}_\mathrm{new}$, while rotation is interpolated using the quaternion slerp operation.

Note that when computing the global pose estimate \eqref{eq:global_pose_calculation}, the current local SLAM result $\mathbf{T}_\mathrm{local}^\mathrm{tracking}$ is used unfiltered, since it is smooth; filtering is only applied to $\mathbf{T}_\mathrm{map}^\mathrm{local}$. This filtering has allowed us to use Cartographer for UAV position control in \cite{milijas2021}. 

\section{Using GPS in aerial SLAM}
\label{sec:GPS}

\begin{figure*}[htbp]
    \centering
    \includegraphics[width=0.4\textwidth]{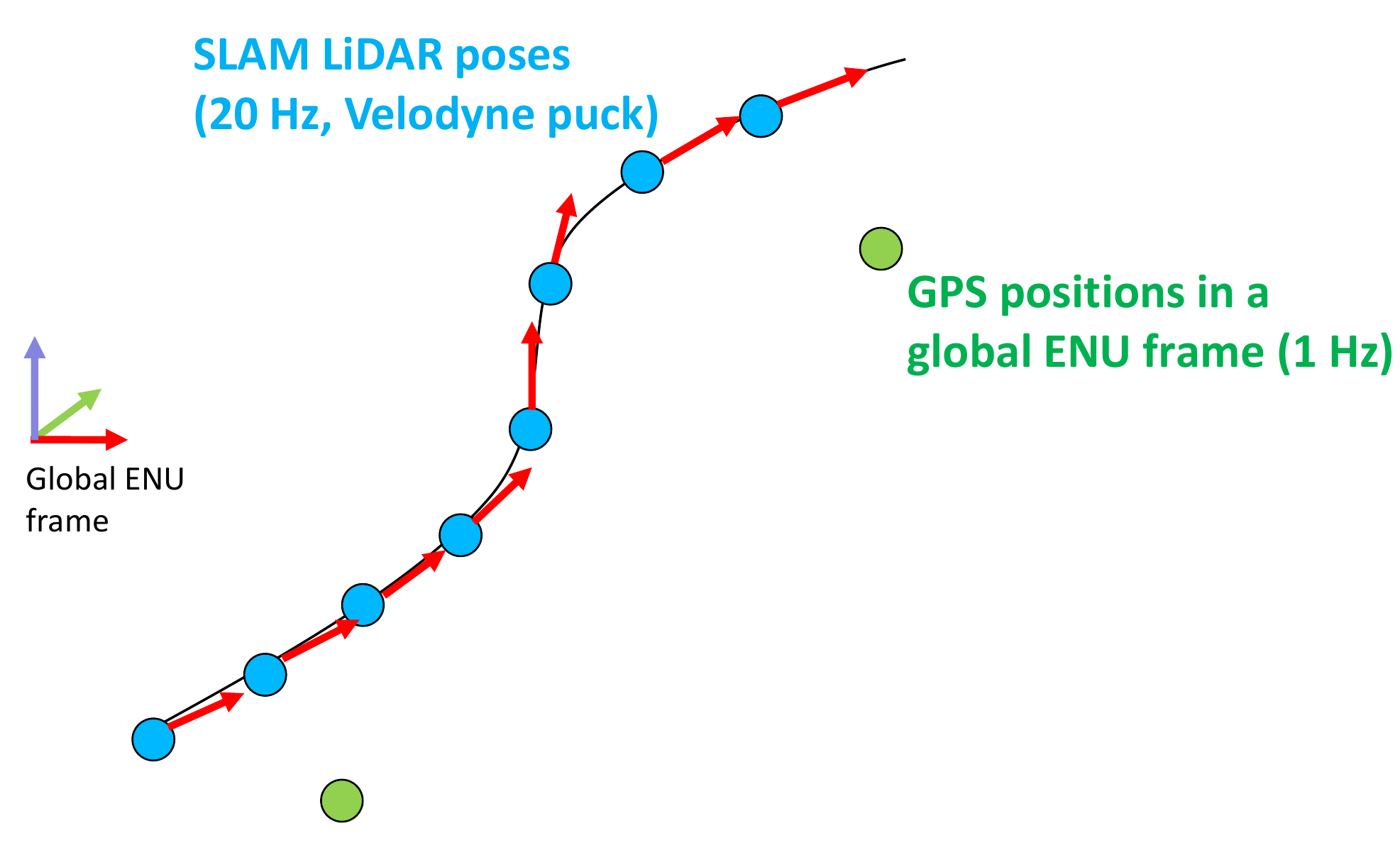}
    \includegraphics[width=0.4\textwidth]{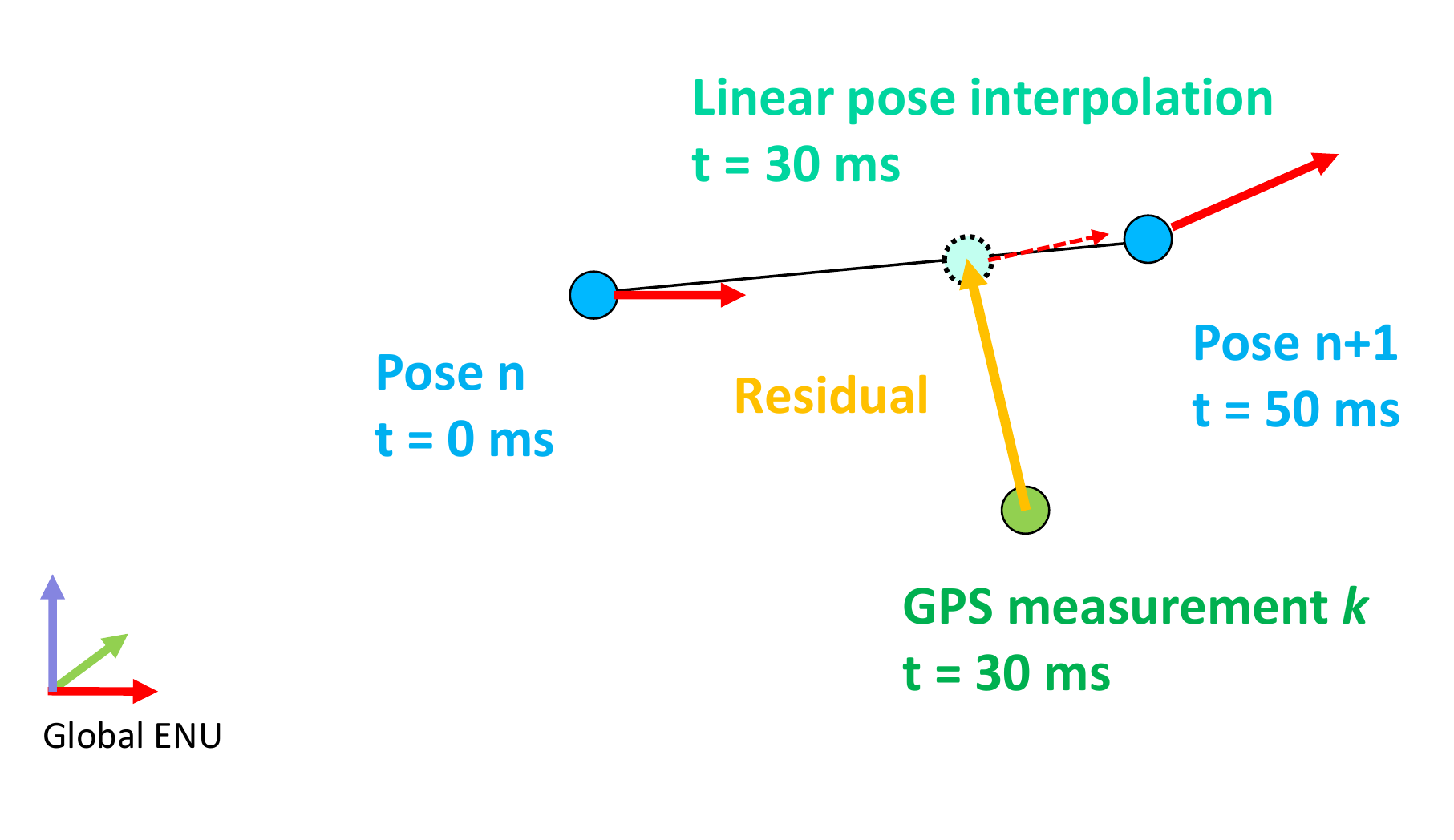}
    \caption{On the left, illustration of the SLAM trajectory poses estimated by proprioception (LiDAR + IMU) and the GPS position measurements. The frequency of GPS measurements is usually an order of magnitude lower than the frequency of LiDAR poses, and there may be intermittent gap intervals of missing measurements in case of GPS signal loss. On the right, illustration of the cost function in the pose graph for constraining the trajectory to a GPS measurement. The residual $\in \mathbb{R}^3$ is computed as the difference of the trajectory position (linearly interpolated to the GPS measurement time), and the position of the GPS measurement. The residuals can be assigned a weight according to the GPS measurement certainty, allowing us to reduce the influence of uncertain measurements.}
    \label{fig:gps_constraint}
\end{figure*}

At the time of implementing, the upstream version of Cartographer did not support using GPS measurements in 3D SLAM (only 2D SLAM supported this with the \emph{fixed frame pose sensor} feature).

We hold that the fixed frame poses feature of Cartographer (which has been upgraded to support 3D SLAM in the meantime) has several drawbacks. The most pronounced drawback is that it is designed for dense (high frequency) poses from an inertial navigation system, due to the fact that fixed frame pose measurements are interpolated to times of LiDAR trajectory poses. Interpolating GPS position measurements in our use case is undesirable, because they are quite sparse (1~Hz or less, in case of signal loss).

Futhermore, in the upstream implementation, the origin of the global (\emph{map}) coordinate system is fixed to the first trajectory pose. We prefer to decouple the trajectory from the map origin and fit it through GPS measurements, while having an east-north-up (ENU) global map coordinate system placed on the surface of the WGS84 ellipsoid. This enables defining mission waypoints using GPS coordinates in the global coordinate system.

\subsection{Residual for GPS measurements}

Figure \ref{fig:gps_constraint} illustrates the reverse principle to the upstream implementation of fixed frame measurements: the dense LiDAR trajectory is interpolated to times of GPS measurements, and not the other way around. This also reduces the complexity of the logic for handling GPS signal gaps -- if there are no GPS measurements, there will simply be no cost functions.

The residual of a constraint  for a GPS position measurement $\mathbf{p}_{\mathrm{GPS,\;}k}$, illustrated in Figure \ref{fig:gps_constraint}, expressed in the global ENU frame is:
\begin{align}
    \label{eq:gps_constraint}
    \bm{res}_{\mathrm{GPS,\;}k}(&\mathbf{T}_\mathrm{map}^{\mathrm{tracking,\;}n}, \mathbf{T}_\mathrm{map}^{\mathrm{tracking,\;}n+1}) \nonumber \\
     = \; & \mathrm{position}(\mathrm{slerp}(\mathbf{T}_\mathrm{map}^{\mathrm{tracking,\;}n}, \mathbf{T}_\mathrm{map}^{\mathrm{tracking,\;}n+1}, \beta)) \nonumber \\
    & - \mathbf{p}_{\mathrm{GPS,\;}k} \nonumber \\
    \triangleq \; & \begin{bmatrix}res_{\mathrm{ENU,}\;x} & res_{\mathrm{ENU,}\;y} & res_{\mathrm{ENU,}\;z}
    \end{bmatrix}^T
\end{align}
where SLAM poses $n$ and $n+1$ with timestamps $t_n$ and $t_{n+1}$ are temporally adjacent to the GPS position measurement $k$ with timestamp $t_{\mathrm{GPS},\; k}$, and the interpolation factor $\beta = \frac{t_{\mathrm{GPS},k} - t_n}{t_{n+1} - t_n}$.

\subsection{GPS measurement weighting}

Precision of the GPS position measurements depends on the signal quality, which can be negatively influenced by a low number of visible satellites, multipath reflections, buildings or foliage. 

Measurement quality is given as the longitude/altitude/latitude standard deviations. Per \cite{Glira:2016:0099-1112:945}, the inverse of the deviations can be used as the weight of the GPS measurement constraint residual. More precisely, if the GPS measurement deviations are given as
\begin{align}
\bm{\sigma}_{\mathrm{GPS,}\;k} = \begin{bmatrix}
\sigma_\mathrm{east} & \sigma_\mathrm{north} & \sigma_\mathrm{up}
\end{bmatrix}^T,
\end{align}
the total weighted cost expressed can be computed as
\begin{align}
    \label{eq:gps_weighting}
    \bm{cost}_{\mathrm{GPS,\;}k} &= \mathrm{diag}(\bm{\sigma}_{\mathrm{GPS,}\;k})^{-1}
    \; \bm{res}_{\mathrm{GPS,\;}k} \nonumber \\
    & = \begin{bmatrix}\frac{res_{\mathrm{ENU,}\;x}}{\sigma_\mathrm{east}} & \frac{res_{\mathrm{ENU,}\;y}}{\sigma_\mathrm{north}} & \frac{res_{\mathrm{ENU,}\;z}}{\sigma_\mathrm{up}}
    \end{bmatrix}.
\end{align}
The role of the inverse weighting scheme is to give more weight in pose graph optimisation to certain (accurate) GPS position measurements, and to reduce the influence of uncertain (noisy) measurements.  

However, in situations where there is a difference between the latitude and longitude certainties, i.e. the east- and north-bound deviations, it has been observed that weighting from \eqref{eq:gps_weighting} can induce an unwanted global trajectory pitch/roll. This can be mitigated by using isotropic weighting for all directions:
\begin{align}
\bm{\sigma}_{\mathrm{GPS,}\;k} = \begin{bmatrix}
\sigma & \sigma & \sigma
\end{bmatrix}^T,
\end{align}
where $\sigma = \mathrm{max}(\sigma_\mathrm{east},\; \sigma_\mathrm{north},\; \sigma_\mathrm{up})$ (the worst of the three), and calculating \eqref{eq:gps_weighting} using the isotropic weight.

Performance of the pose graph optimization process can often be manually improved by adjusting the weights assigned to different constraints in the pose graph. For this reason, \eqref{eq:gps_weighting} can be modified  into an affine function with user-supplied scalar values $a$ and $b$:
\begin{align}
    \label{eq:gps_weighting_ab}
    \bm{cost}_{\mathrm{GPS,\;}k} &= (a \; \mathrm{diag}(\bm{\sigma}_{\mathrm{GPS,}\;k})^{-1} + b \mathbf{I}) \; \bm{res}_{\mathrm{GPS,\;}k}
\end{align}
The user can change the influence of GPS constraints on the rest of the pose graph by appropriately modifying the values $a$ and $b$, where $b$ controls the influence of all GPS measurements uniformly, while $a$ controls the influence of high-certainty GPS measurements (the measurements with low deviation). For example, if the optimizer does not closely align the trajectory with the GPS measurements, the weights can be increased until the desired behaviour is obtained.

For robustification of pose graph optimization, the influence of outliers in the set of GPS measurements can be reduced by applying a loss function (e.g. Huber loss) to the GPS measurement cost. In other words, measurements which are a certain distance away from the trajectory have a suppressed attractive effect on the trajectory in pose graph optimization.

\subsection{Initial orientation}

An additional step towards using georeferenced Cartographer poses for UAV navigation was to set the orientation of the local SLAM starting pose, so that correct vehicle heading is available from the get-go.
This avoids having to perform kinematic alignment of the trajectory heading, i.e. introduction of large residuals into the pose graph which would lead to large shifts in the pose estimate in the early stages of the mission. 

In order to be able to set the vehicle starting orientation, we have modified the \texttt{start\_trajectory} ROS service in Cartographer, which normally allows the user to manually start a new SLAM trajectory with a relative pose with respect to another trajectory. Since we wish to set an absolute initial vehicle orientation (i.e. relative to the origin of the global ENU map frame), we have added a virtual trajectory with an ID of -1 which represents the global ENU map frame origin. When starting a new SLAM trajectory, this allows us to use the UAV orientation information from the UAV autopilot (which in turn is sourced from the magnetometer) as the initial SLAM orientation.


\section{Navigation}
\label{sec:navigation}

Apart from using Cartographer to estimate the global UAV pose, successful UAV navigation requires a way of using the constructed map to plan a collision-free trajectory for the UAV to execute. Cartographer builds the map by stitching together many smaller submaps which are internally stored as \emph{hybrid grids}. Hybrid grids are voxel grid pairs, where \emph{hybrid} refers to the fact that each grid has a low-resolution and high-resolution variant. This is used to speed up the scan matching process: the low-resolution grid is used for coarsely aligning the received scan to a map, while the high-resolution grid is utilised for fine alignment.

To the best of our knowledge, a path planning method suitable for such structures is yet to be developed.  To mitigate this problem, we have devised a system which uses an OctoMap \cite{Hornung2013} to create an octree based map of the environment, suitable for path planning and navigation. 

The OctoMap is a hierarchical volumetric 3D representation of the environment. Each cube of the OctoMap is called a voxel, which can be \textit{free}, \textit{occupied} or \textit{unknown}.

The OctoMap is generated using Cartographer submaps. The submap $M^{i}_{s}$ ($i^{th}$ submap) is an occupancy grid map built using the last $n_s$ consecutive LiDAR scans $S$ which passed motion filtering, and matched with past IMU and odometry data \cite{Batinovic2021}: 
\begin{equation}
M^i_{s} = f(S_{(i-1)n, s}, \ldots, S_{in,\; s}, \; \mathrm{IMU}, \; \mathrm{Odometry}).
\label{eq:submap}
\end{equation}

The function $f$ stands for a nonlinear optimization that aligns each successive scan against a submap being built. A submap is marked as completed when the predetermined fixed number of scans $n_s$ are inserted therein. 
 
The map $M$ can be created by joining all past submaps together:
\begin{equation}
M^i = f(M^1_s, \ldots, M^i_s).
\end{equation}

\begin{figure}[t!]
	\centering
	\includegraphics[width=1\columnwidth]{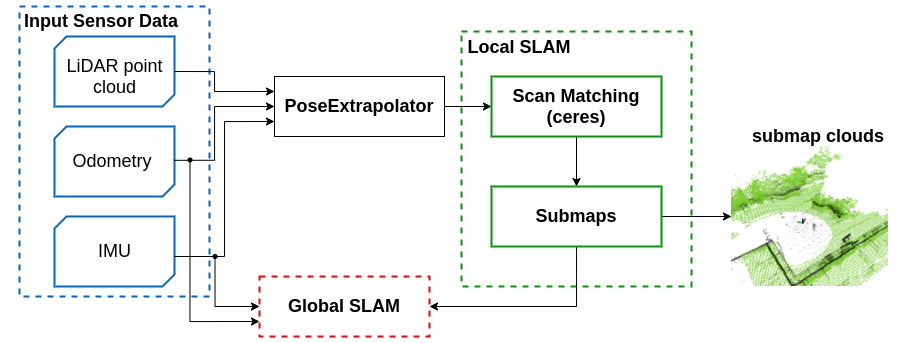}
	\caption{Submap clouds extraction from the submaps module inside Cartographer local SLAM.}
	\label{fig:submap_cloud_diagram}
\end{figure}

Both the map $M$ and the submaps $M_s$ are in the form of a 3D occupancy grid. Since Octrees are a format much more suitable for navigation operations such as path planning, instead of building a 3D occupancy grid map $M$, we build an OctoMap $O$ using the OctoMap generation software \cite{Hornung2013}. Namely, from each completed submap $M^i_s$ we calculate a submap cloud $M^i_{sc}$ by adding a point in the centre of each occupied voxel of $M^i_s$. Figure \ref{fig:submap_cloud_diagram} shows a submap cloud extracted from a submap.

The OctoMap $O^i$ is then created from all the past submap clouds:
\begin{equation}
O^i = f(O^{i-1},\; M^i_{sc}),\; O^0 = \emptyset.
\label{eq:octomap_generation}
\end{equation}

Note from \eqref{eq:submap} that the size of a submap is adjustable, where $n_s$ laser scans are matched to form a submap. Submap clouds provide a more compressed yet comprehensive input for OctoMap generation when compared to raw LiDAR data. The OctoMap is then further used for path planning and safety navigation.

\begin{figure}[t]
	\centering
	\includegraphics[width=1\columnwidth]{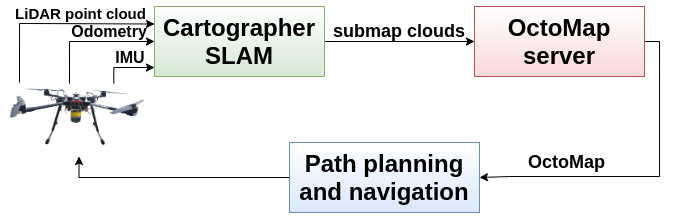}
	\caption{Overall schematic diagram of navigation using an OctoMap. The Cartographer SLAM creates a submap cloud, which is an input for the OctoMap server module. The path is planned in the OctoMap and the robot navigates by closing the loop with the Cartographer SLAM module.}
	\label{fig:navigation_diagram}
\end{figure}

An overview of the proposed system is given in Figure \ref{fig:navigation_diagram}. The Cartographer SLAM algorithm requires an appropriate sensing system, e.g. a laser scanner or a camera to create a 3D map. An OctoMap is generated using the SLAM algorithm (submap clouds) and is used for collision-free navigation. 


Cartographer submap hybrid grids are dual-resolution voxel occupancy grids, as discussed previously. For noise reduction, during the conversion to OctoMap, we have opted to use the lower resolution occupancy grid, as well as a 0.7 voxel occupancy probability threshold (i.e. a voxel is included in the submap cloud passed to OctoMap only if its occupancy probability exceeds 0.7).

For UAV control, we use an RRT-based path planner and trajectory following solution \cite{Arbanas2018}. The planner avoids occupied voxels from the OctoMap and generates a path through the free voxels.

\begin{figure*}[t]
	\centering
	\includegraphics[width=0.75\textwidth]{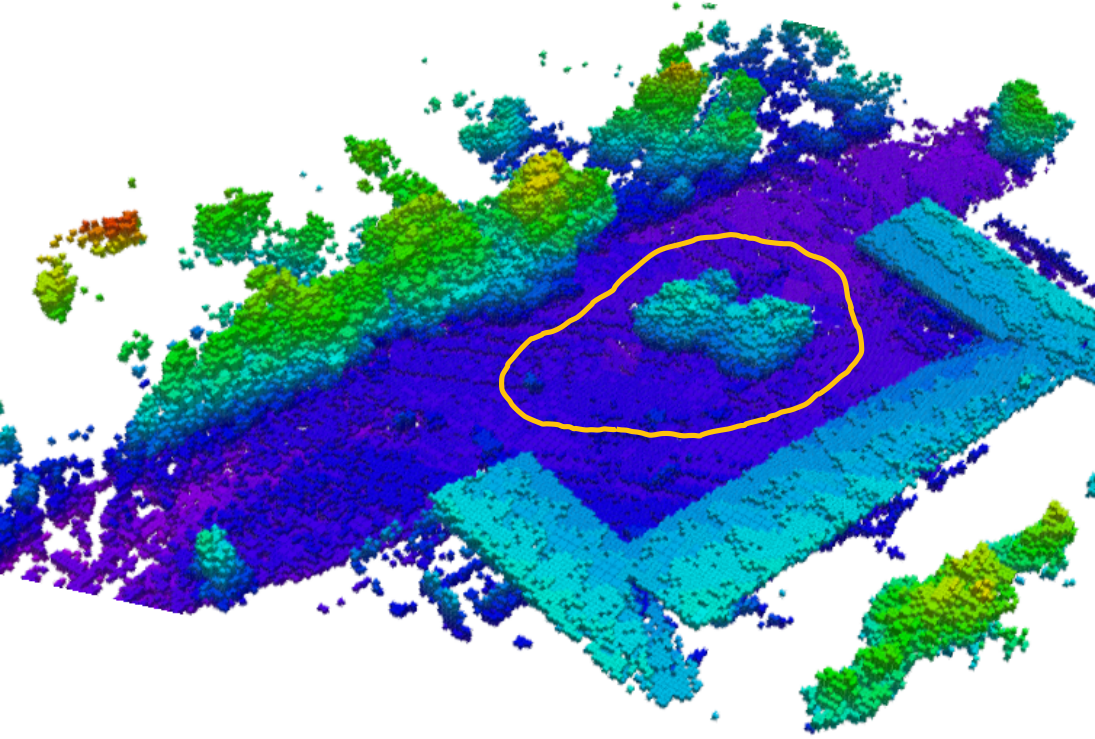}
	\caption{The OctoMap created during the navigation with the path traversed by the UAV.}
	\label{fig:octomap}
\end{figure*}

The SLAM algorithm plays an important role in many applications, such as civil infrastructure surveys, search and rescue, and autonomous exploration. Cartographer provides a 
map suitable for both 2D \cite{Orsulic2019}, \cite{Batinovic2020} and 3D \cite{Batinovic2021} autonomous exploration, employing the well known frontier detection method for exploration planning and execution.
\section{Experiments discussion}
\label{sec:experiments}

\subsection{Hardware description}

\begin{figure}
    \centering
    \includegraphics[width=\columnwidth]{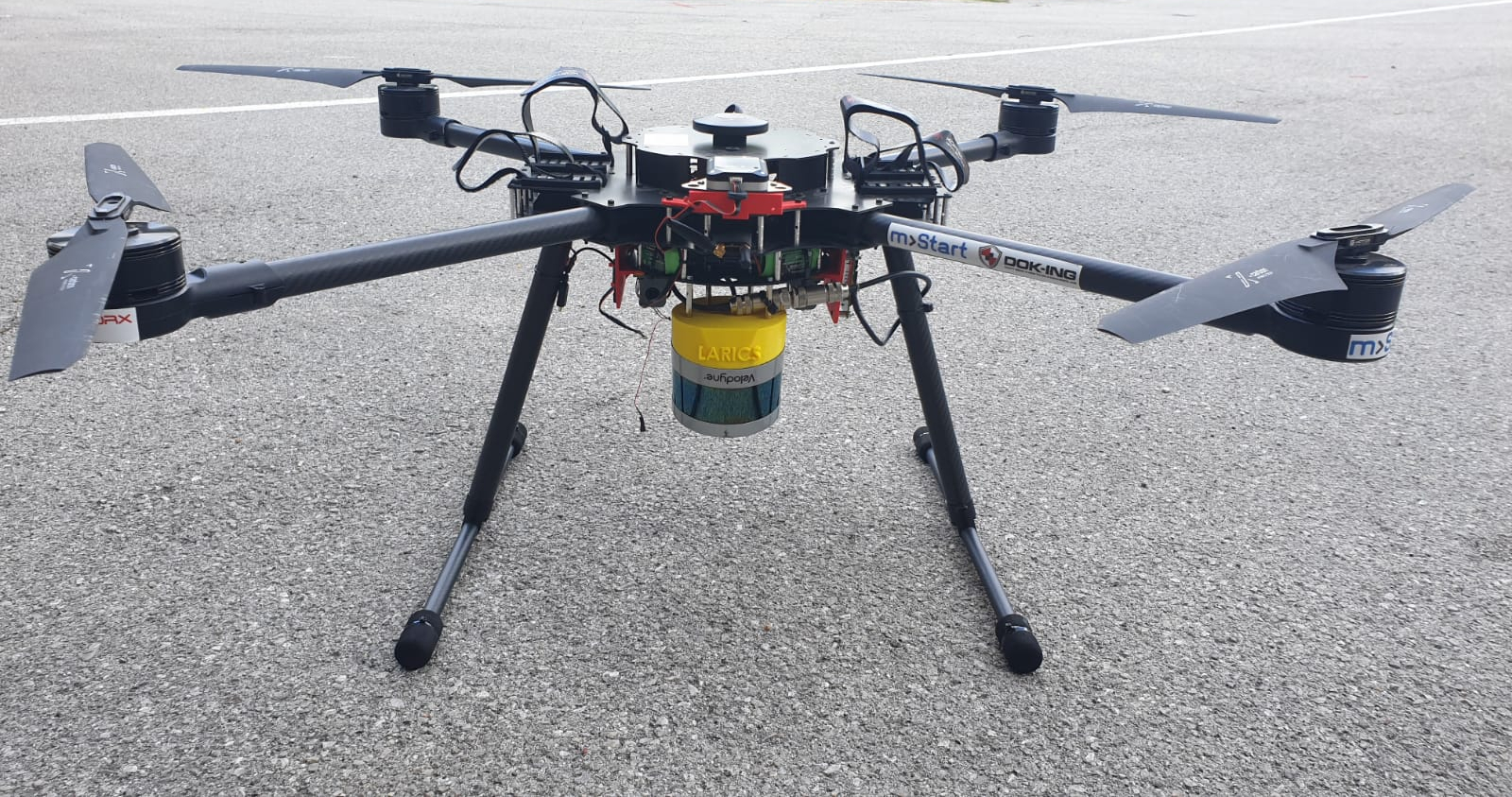}
    \caption{The aerial platform used for our experiments.}
    \label{fig:kopter}
\end{figure}

Our experimental setup includes a custom-built quadcopter assembled by the company \textit{Kopterworx}, shown in Figure \ref{fig:kopter}. It features four T-motor P60 KV170 propulsion units with $22''$ folding propellers. The dimensions of the vehicle are $1.2 \; \mathrm{m} \times 1.2  \;  \mathrm{m} \times 0.45  \;  \mathrm{m}$ with mass $ \mathrm{m}   =  9  \; \mathrm{kg}$, including all electronics and batteries. The larger vehicle scale enables a larger on-board battery capacity, allowing for $30 \; \mathrm{min}$ of autonomous flight time. We use the Pixhawk 2.1 flight controller with a ProfiCNC/HEX Kore high-current power board. We also mounted an Intel NUC onboard computer that communicates with the flght controller through serial communication. The NUC is accessible to the operator through a WiFi interface. The operating system installed on the on-board computer is Ubuntu Linux 18.04 LTS with ROS Melodic middleware. Apart from the flight controller, the onboard computer communicates with the complete sensory apparatus of the vehicle and runs high-level computationally intensive algorithms. Furthermore, we attached a Velodyne Puck LITE LiDAR, LPMS CU2 external IMU and HERE+ GPS based on the U-blox M8P IC.
The experiments were conducted outdoors on the Borongaj university campus. GPS measurements were used for georeferencing the trajectory in the experiment shown in Figure \ref{fig:GPS comparison}.

\subsection{Safe loop closures}

The graph from Figure \ref{fig:steps_in_feedback} was recorded on a circular trajectory around a small building. The trajectory was long enough for the global SLAM to insert some loop closure corrections into the pose graph and use them in optimization. These are clearly visible on the z-axis plot shown in Figure \ref{fig:steps_in_feedback} since they introduce a step disturbance into the pose estimate. The scaling factor introduced with \eqref{eq:smooth} eliminates these step disturbances and ensures a smooth pose estimate.

\subsection{Using GPS in aerial SLAM}

Since most UAVs come equipped with GPS sensors by default, we have attempted to utilise this sensor to georeference the SLAM trajectory. On our testing site, GPS usually has a standard deviation larger than 1.5 meters and as such it has proven to be challenging to use in trajectory optimization. The position signal filtered by Pixhawk has pronounced vertical drift -- there is a difference of 80 cm between the takeoff and landing altitude measurements (which should, in fact, be the same).

\begin{figure}
    \centering
    \includegraphics[width=1.0 \columnwidth]{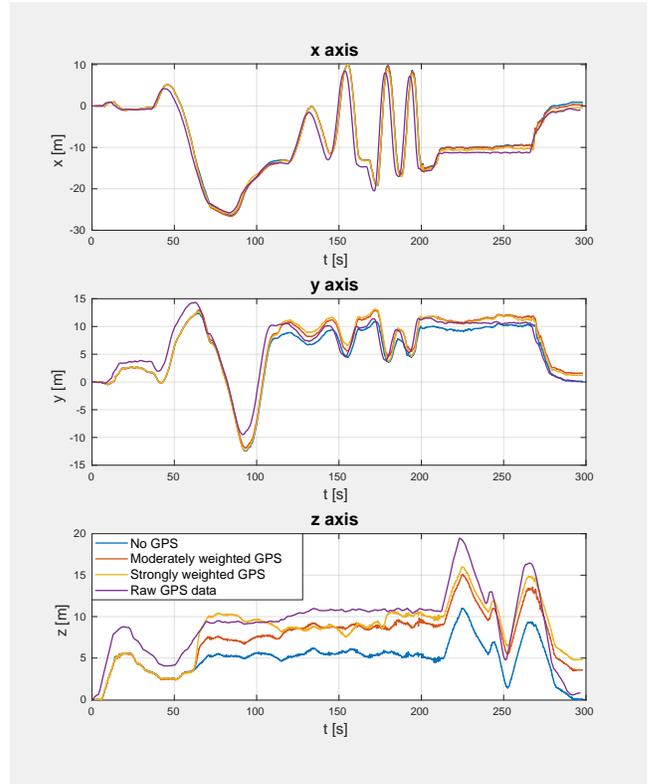}
    \caption{Comparison of UAV coordinates reported by Cartographer for different GPS settings. The GPS data as reported by Pixhawk is also shown. It is visible that the GPS altitude differs significantly from the altitude reported by Cartographer without GPS, especially during takeoff. This has a negative effect on the pose estimate when GPS measurements are used for optimization, causing Cartographer to report a non-zero landing altitude. Since GPS measurements are only taken into account during periodical pose graph optimizations, inaccurate GPS measurements can cause failures to close loops, and disturbances in the pose estimate (which can be mitigated by robustification using a loss function, adjusting the weights \eqref{eq:gps_weighting_ab} or filtering, as described in section \ref{sec:loops}). The first optimization is particularly aggressive, since it tries to reduce the gap between the GPS trajectory and the SLAM trajectory which is greatest at this time. The frequency of pose graph optimizations is controlled by the user -- in our case, optimizations are performed once per minute.}
    \label{fig:GPS comparison}
\end{figure}

We have attempted to tune the weight which controls how the GPS measurements influence the pose estimate, but we have not managed to entirely eliminate the negative effect inaccurate GPS measurements have on the pose estimate. Figure \ref{fig:GPS comparison} shows the pose estimate reported by Cartographer without using GPS measurements, and with two different GPS measurement weights. The GPS measurements reported by the Pixhawk (and thus a trajectory constrained to them) are inconsistent, since the reported takeoff and landing altitudes are in mismatch. The figure also shows an aggressive change in the reported altitude at the first pose graph optimization.

A short SLAM trajectory -- such as the one in Figure \ref{fig:GPS comparison} processed without GPS measurements -- can be considered as ground truth, albeit not georeferenced, since distinct LiDAR point cloud features such as walls and roof edges remain consistent within the map. This consistency is primarily due to the fact that laser scans in different parts of the trajectory are cross-registered correctly, thanks to the discovered loop closure constraints.

Use of inaccurate GPS measurement constraints can overpower the loop closure constraints which register the trajectory to itself and cause object separation in the built map. If this is noticed, it is advisable to reduce the weight of GPS measurements.


\subsection{Navigation}

Extracting Cartographer submaps, converting them into PointClouds and using those to build an OctoMap, allows our UAV to plan collision-free paths in its environment. Therefore, the introduction of OctoMap in our system allows us to use Cartographer not only for pose feedback but also as a source of information about the environment useful for navigation.

An example of an OctoMap built using Cartographer submap clouds is shown in  Figure \ref{fig:octomap}, in which the path traversed by the UAV during the navigation is also shown. 

\section{Conclusion}
\label{sec:conclusion}

After presenting the changes we have made to Cartographer in order to adapt it for UAV navigation, we conclude that filtering of loop closures and the extraction of submap clouds are helpful additions to  th Cartographer code base. The former makes the pose estimate more suitable for control system feedback as it removes the step signals which are introduced during loop closure events, while the latter creates the possibility of using  submaps for planing collision free paths for the UAV to execute. 


A possible direction of future work is tighter and more efficient integration of the navigation and planning layer (OctoMap/RRT path planner) with the pose graph, taking advantage of its structure (immutable submaps being displaced by pose graph optimization), while avoiding OctoMap agglutination caused by displacement of submaps.

It can also be concluded that it is currently best to avoid the use of the M8P GPS sensor in this particular configuration with Cartographer due to its detrimental effect on the pose estimate quality. The described GPS alignment methodology was previously successfully tested and used in a control loop on a ground vehicle (the Husky AGV), and also with a proprietary commercial survey-grade LiDAR data capture system based on an inertial navigation system made by the manufacturer \emph{Novatel}. We hope to show similar success in our future work with UAVs, employing other GPS configurations, e.g. use of another GPS receiver, a base station, and RTK or DGPS positioning modes.

\section*{ACKNOWLEDGMENT}

This work has been supported by the European Commission Horizon 2020 Programme through project under G. A. number 810321, named Twinning coordination action for spreading excellence in Aerial Robotics - AeRoTwin \cite{AEROTWINweb} and through project under G. A. number 820434, named ENergy aware BIM Cloud Platform in a COst-effective Building REnovation Context - ENCORE \cite{ENCOREweb}.
\bibliographystyle{ieeetr}
\bibliography{bibliography.bib}

\vspace{12pt}

\end{document}